# Attention-Mechanism-based Tracking Method for Intelligent Internet of Vehicles

Xu Kang[1], Bin Song[1], Jie Guo[1], Xiaojiang Du[2] and Mohsen Guizani[3]

[1]State Key Laboratory of Integrated Services Networks, Xidian University, Xi'an, P.R. China
[2]Department. of Computer and Information Sciences, Temple University, Philadelphia, USA
[3]Department. of Electrical and Computer Engineering, University of Idaho, Moscow, Idaho, USA

**Abstract**

Vehicle tracking task plays an important role on the internet of vehicles and intelligent transportation system. Beyond the traditional GPS sensor, the image sensor can capture different kinds of vehicles, analyze their driving situation and can interact with them. Aiming at the problem that the traditional convolutional neural network is vulnerable to background interference, this paper proposes vehicle tracking method based on human attention mechanism for self-selection of deep features with an inter-channel fully connected layer. It mainly includes the following contents: 1) A fully convolutional neural network fused attention mechanism with the selection of the deep features for convolution. 2) A separation method for template and semantic background region to separate target vehicles from the background in the initial frame adaptively. 3) A two-stage method for model training using our traffic dataset. The experimental results show that the proposed method improves the tracking accuracy without an increase in tracking time. Meanwhile, it strengthens the robustness of algorithm under the condition of the complex background region. The success rate of the proposed method in overall traffic datasets is higher than Siamese network by about 10 percent, and the overall precision is higher than Siamese network by 8 percent.

**Keywords**

Vehicle tracking, bilinear network, attention mechanism, semantics, response

## Introduction

The internet of vehicles (IOV) can improve people's travel efficiency through urban traffic management, traffic congestion detection, path planning, road charge and public transportation management, to alleviate traffic congestion. By the surveillance camera in the bayonet and key sections of the city, we can perform recognition and tracking of all types of vehicles. Statistics on the server through the identification and tracking results, we can calculate the driving path and determine the intention of the moving vehicle. In this way, we can also analyze the real-time road conditions at the location of the sensors, and then feed the results back to the user's vehicle through a wireless sensor, guiding the next step and recommending the appropriate route. Vehicle tracking is a key technology of IOV and intelligent transportation system (ITS), in which the image sensor and wireless sensor are complementary. The accuracy and speed of vehicle tracking system directly affect the performance of IOV. In recent years, with the development of computer hardware and improving the performance of intelligent algorithms, the performance of the vehicle tracking system is also increasing. The goal of vehicle tracking task is to get the position information of the target vehicle in the first frame in a video or an image sequence. In each subsequent frame, the position of the vehicle is predicted by various operations, including the center coordinates and the width and height of the target vehicle. The difficulty of vehicle tracking is how to select effective feature extraction methods for different scenes to express the target image region, so that the tracking model can effectively learn and predict the input samples.

Since 2012, Convolution Neural Network (CNN) has been popularized in the field of computer vision. More and more researchers in the visual field choose CNN as the main body of their algorithm. Convolution neural network compared with the traditional pattern recognition method, the biggest advantage is to save the previous heavy feature engineering process. In natural scenes, illumination and color are different. It is difficult to find a universal feature extraction method to apply to all scenes. The convolution neural network can transform the feature of the pixel domain into the high-level abstract feature through the convolution operation of the image, and the parameters used in the extraction process can be trained and learned by a large amount of data, and the interference of human factors is eliminated.

In [1]-[4], the authors try to use correlation filter of the target area on the feature map of the convolution neural network. They try the different feature layers and different structure neural networks, but the accuracy of the tracking is not improved, and the time delay of each frame is greatly increased. In [5], Nam and others discard a large scale convolution neural network, considering about a simple image block, using a small network of VGG-M (only three layers of coiling layer) to extract the image blocks around the target to determine whether it is a target object. This model has a very considerable accuracy rate, but it needs to extract a large number of candidate regions around the target area for decision. Each area needs to be extracted by the neural network, and the full connection layer of the network needs to be updated online. The tracking speed can only reach 1 frames per second (FPS). In [6][7], the authors use two branch neural networks to calculate the similarity between two image blocks. When tracking, the target image block in the last frame and the surrounding image block are sent to two convolution neural networks respectively, and then the output results are sent to the same full connection layer to judge the similarity degree of the two. This method has omitted the online update process, and after one training, the network only forecasts each frame without two training, and improves the time delay and accuracy. In [8][9], the authors use the fully convolutional neural network to measure the similarity between the two frames, which is to extract the convolution neural networks from the two frames with the same parameters. The output results do not pass through the fully connected layer but are convoluted with each other to obtain the similarity feature map, so that the coordinates of the target rectangle frame in the entire image frame are obtained.

In this article, we propose a fully convolutional neural network combined with an attention mechanism to measure the similarity of a template and a search area using selected deep features of multiple channels. We separate template and semantics from the background area of target vehicles in the initial frame adaptively. We use a two-stage method and an analog loss function for model training. The experimental results show that the proposed method achieves ideal accuracy with a decrease in tracking time. The robustness of algorithm under the condition of complex background region has been strengthened. The proposed method in overall traffic datasets on the success rate is higher than KCF, DSST and Siamese-FC by at most 15 percent, the overall precision is higher than the three methods by at most 20 percent.

The rest of the article is organized as follows: Section "Background and related work'' introduces the traditional CNN and similarity neural networks. Section ''Semantic Attentional Bilinear network (SAS-Net)'' describes the semantic attention similarity neural network for vehicle tracking and implementation of it, including the architecture of our similarity neural network, the architecture of our attention mechanism, and an extraction method of adaptive target scale. The loss function and a two-stage training method are presented in section ''Loss function and training.'' Section ''Experiments'' compares the success rate, precisions and speed of the proposed SAS-Net with other three methods based on the experiments on our traffic dataset. We conclude this article and discuss the next research focus in section ''Conclusion and future work.''

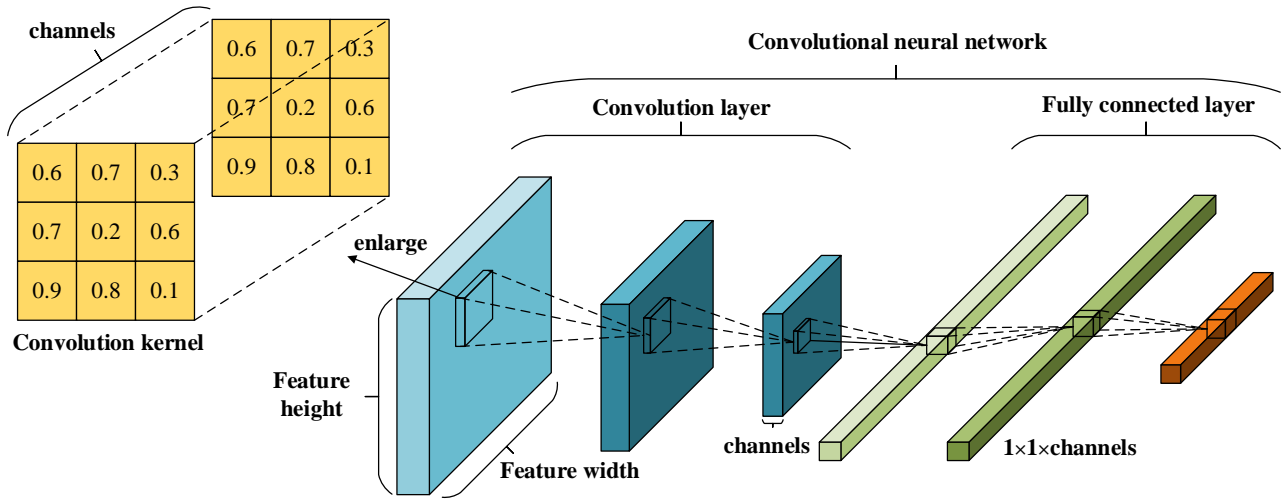

**Figure 1.** Architecture of a convolutional neural network. The general convolution neural network is mainly composed of a convolution layer and a fully connected layer, usually with a down-sampling layer and an activation function layer. Each neuron in the convolution layer is connected only to the part of the previous feature map. For the fully connected layer, it is to connect all the output of the upper neuron and output a one-dimensional feature vector.

## Background and related work

### *convolutional neural networks*

The general convolution neural network, as shown in Figure 1, is mainly composed of a convolution layer and a fully connected layer, usually with a down-sampling layer and an activation function layer. Each neuron in the convolution layer is connected only to the part of the previous feature map. The number of parameters is the size of the convolution kernel and the kernels are shared, which greatly reduces the weight of the convolution. If the step size is increased, the overlap area can be reduced when the convolution kernel slips, and the number of operations can be reduced. For the fully connected layer, it is to connect all the output of the upper neuron and output a one-dimensional feature vector to play a unified role in dimension. Since the full connection layer is connected to all the output of the previous layer, the number of parameters it contains is much more than the convolution layer, and there is lots of redundancy, so it needs to be used with some specific activation functions. For example, the dropout layer can inhibit neurons from propagating forward with a certain probability (usually 0.5). The down-sampling layer samples the original feature map to a smaller size without any multiplication or addition operation without any parameters. The sliding method of the lower sampling area in the lower sampling layer is the same as the method of the convolution kernel sliding in the convolution layer. When the current sampling area slides to a position, the maximum value or the average value of all the pixels in the sampled area is taken instead of the whole region. This process is separately called the maximum pooling and the average pooling process. The pooling process is the best method to reduce the amount of data of the feature map, which does not change the feature structure of the image as a whole, and can be intuitively understood as the reduction of the resolution of the feature map.

### *Similarity neural network*

The similarity neural network is a little different from the traditional CNN. The two image blocks, through the convolution layer and pooling layer, output the high dimensional feature maps. Then the features are input into the decision network to calculate the similarity of two image blocks[11]. The decision network is usually a fully connected neural network. The output result is a similarity value, representing the matching degree of the two image blocks. Before training, the matching degree of two image blocks needs to be labeled artificially. If the two image blocks match, the output label value is labeled as y=1. If the two image blocks do not match, then the label of the training data is labeled as y=0. The annotation method of training data is not a similarity value at all, but a boolean value. The loss function of the convolutional neural network is the cross-entropy between the true probability distribution of the input image and the probability distribution predicted by the network. It is the sum of the entropy of label distribution and the KL divergence between label distribution and predicted distribution:

$$\begin{aligned} H(p,q) &= -\sum_x p(x)\log q(x) \\ &= -\sum_x p(x)\log p(x) - \sum_x p(x)\log\frac{q(x)}{p(x)} \\ &= H(p) + KL(p||q) \end{aligned} \quad (1)$$

It represents the information difference between two probability distributions, the greater the value, the greater the difference between the two probability distributions, which means that the two probability distributions are closer. The illustration of the bilinear similarity neural network is shown in Figure 2. The main idea is to pass the two image blocks through two branches, respectively, to input the subsequent two features into a fully connected decision network. If the two branches have completely different network structures and weight values, the feature extraction functions of the two branches are unrelated and different. The two extracted features differ in values and dimensions. When the two high-level feature maps pass through the first fully connected layer, their feature vectors are fused together. The parameters of this network are nearly twice as much as the bilinear network with shared weights. Also, it requires a greater amount of data and longer time in training, but it is more flexible.

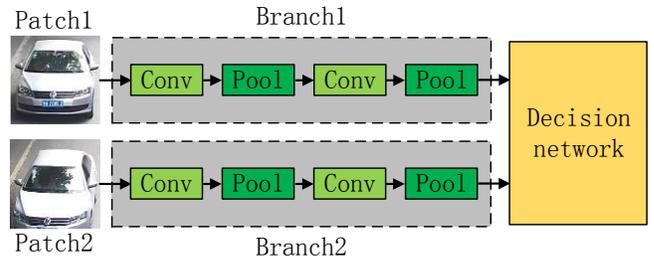

**Figure 2.** Bilinear similarity network. The two image blocks are passed through two branches, respectively. Then the subsequent two features are input into a fully connected decision network.

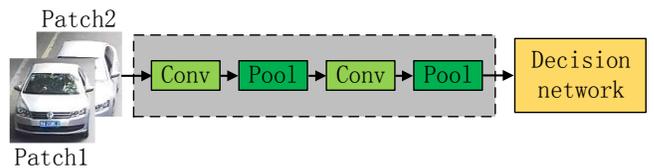

**Figure 3.** Double channel bilinear network. The two image blocks are superimposed as the input. They share both the weights and feature maps during forwarding propagation.

Figure 3 shows the illustration of the dual channel similarity neural network. The two image blocks are superimposed as the input. They share both the weights and feature maps during forwarding propagation. The scale of the model is greatly reduced.

For a long time, there are many algorithms using similarity neural network before the vehicle tracking task. As early as 2005, Chopra et al. used a bilinear structure in [10] to measure the face similarity. In [12], Lin et al. use a similar structure of the bilinear network in fine-grained bird classifications, which uses one branch to locate the object, and the other one to extract the features. Bertinetto et al. use the full convolutional Siamese network in the [9] to search the closet area to the original target in the surrounding area of the target object, and the online learning is not needed in this process.

## Semantic Attentional Bilinear network (SAS-Net)

In recent years, visual attention mechanism 错误!未找到引用源。[22] is one of the most popular research hotspots on visual images. The models based on CNN and recurrent neural network (RNN) is widely used in various types of visual tasks, such as target recognition, fine-grained classification, image segmentation, image caption and visual question answering (VQA). In these tasks, common usage of attention mechanisms is to extract features from text or images, and embed them into another CNN to achieve information fusion. The advantage of the attention mechanism is its ability to predict the weight vector for feature maps by model learning.

A typical example is in [17], Seo P et al. fuse the digit and their color feature vectors in the image, making the network predict the color of the digit while focusing on the location of the digit. Inspired by this model, the tracking model of fully convolutional neural network embedded by an inter-channel visual attention mechanism is proposed in this paper. With the target area combined with its background pixels as the semantic input in the first frame, the model can selectively predict the channel weights of convoluted template features and search area features according to feature maps of different channels. The weights are embedded into the corresponding channels of convoluted features to enhance or weaken the features of some channels.

In the CNN, the high-level features of semantics are robust to the changes of the target appearance, which improves the discriminative power and enhances the generalization ability of the model. In order to make the model more discriminant, we use the inter-channel attention mechanism. Intuitively, the contributions of each channel to the predicted similarity value are different. For the vehicle tracking task, the features of some channels seem to be more important and the others may have no significant impact on the predicted results. If the model can calculate the similarity according to features of different channels selectively, it will improve the performance of the model for vehicle tracking task.

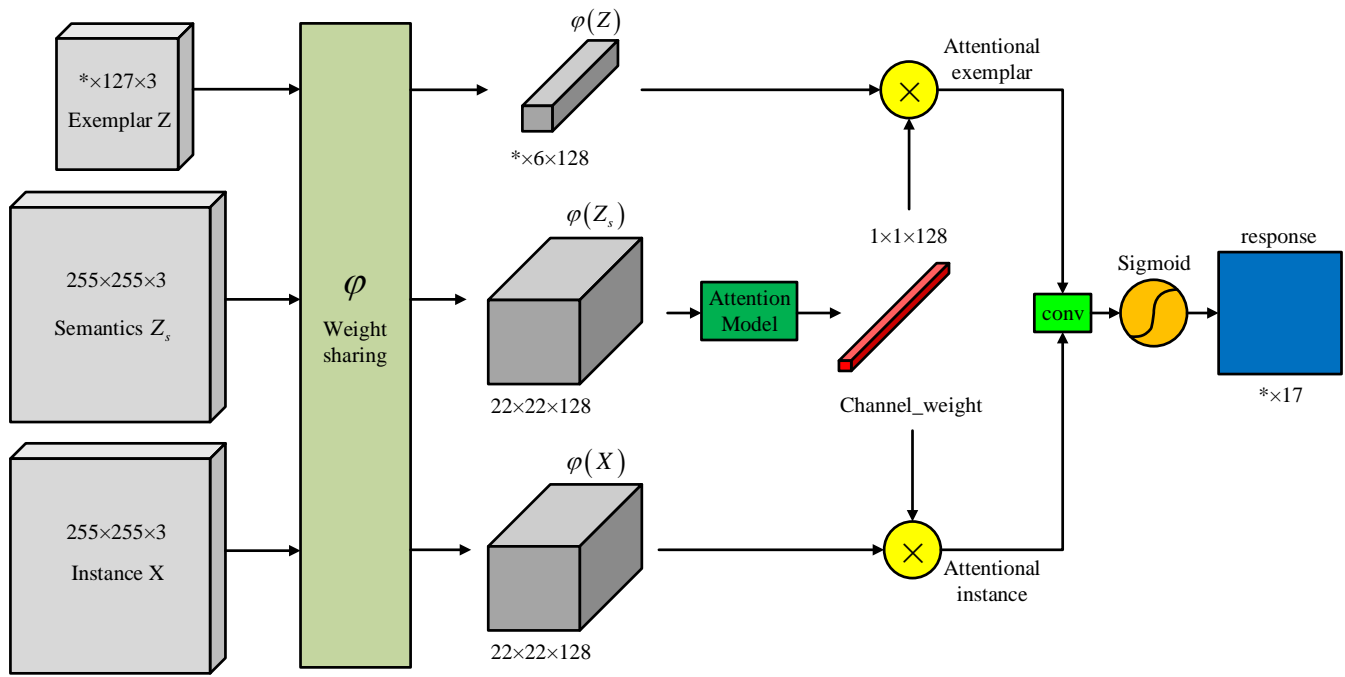

**Figure 4.** Architecture of SAS-Net. The exemplar $Z$, the semantics $Z_s$ and the instance $X$ are all passed through the fully convolutional neural network with shared parameters. The corresponding feature maps of each input image

are $\varphi(Z)$, $(Z_s)$, $\varphi(X)$. $\varphi(Z_s)$ passes through the attention mechanism and get a channel vector. It used to compute the attentional exemplar and attentional instance. The final response is the convolution of the two features. The details of attention model is shown in Figure 5.

## Architecture of SAS-Net

The structure of proposed SAS-Net is shown in Figure 4. Unlike previous networks that require a fixed size image as input, a template $Z$ based on the adaptive length-width ratio is used as a vehicle sample, in order to make the background information in vehicle features as little as possible. When a video from an image sensor input, the image patch of the vehicle in the first frame is extracted. Its long side is resized to 127 pixels, and the whole vehicle patch is resized according to the same ratio as an exemplary image.

The template in the first frame surrounded by the background pixels is extracted as a square patch. Then it is resized to size $255 \times 255$ and is input into the network as a semantic image. In each subsequent frame of the video, the candidate area predicted from the previous frame is resized to size $255 \times 255$ as an instance image $X$. The exemplar $Z$, the semantics $Z_s$ and the instance $X$ are all passed through the fully convolutional neural network with shared parameters, respectively obtain the corresponding feature maps of each input image $\varphi(Z)$, $(Z_s)$, $\varphi(X)$. Among them $\varphi(Z_s)$ and $\varphi(X)$ have the same size. $\varphi(Z_s)$ passes through the attention mechanism and get a channel vector of 128 pixels. It means the weight distribution of the semantic features obtained by the information of exemplar combined with the information of background area. Then the exemplary feature $\varphi(Z)$ and the instance feature $\varphi(X)$ are element-wisely multiplied by the 128-channel weights respectively. The features of each channel are filtered according to the different importance, and the feature of the exemplar and the instance after the multiplication sampling are obtained. Compared to the feature $\varphi(Z)$ and $\varphi(X)$ in the bilinear network, the semantic information is added to the network, which makes it clear which channels are significant and which channels are unnecessary when computing the similarity. Using the same convolution operation, the similarity response map can be calculated by the convolution of the two features filtered by the attention mechanism.

The final similarity response map is output by mapping all the values of the similarity response to 0 to 1 through the sigmoid function. The whole process can be expressed in formula (2), where $Z_s$, $Z$, $X$ respectively represent semantic images, templates and search areas. $\varphi(Z_s)$, $\varphi(Z)$, $\varphi(X)$ represent the feature maps of the three input images through the same neural network $\varphi$ with shared parameters, and $\theta_s, \theta_{att}$ represent the parameters of the bilinear network and the parameter of the attention mechanism respectively. $att(\cdot)$ indicates the mapping function of the attention mechanism, and the $\odot$ indicates the element-wise multiplication of two feature maps, and the $R^{att}$ represents the response of the final output of the SAS-Net model. The proposed model finally maps the similarity response map to 0 to 1 through a sigmoid function, so that all values on the response map are positive.

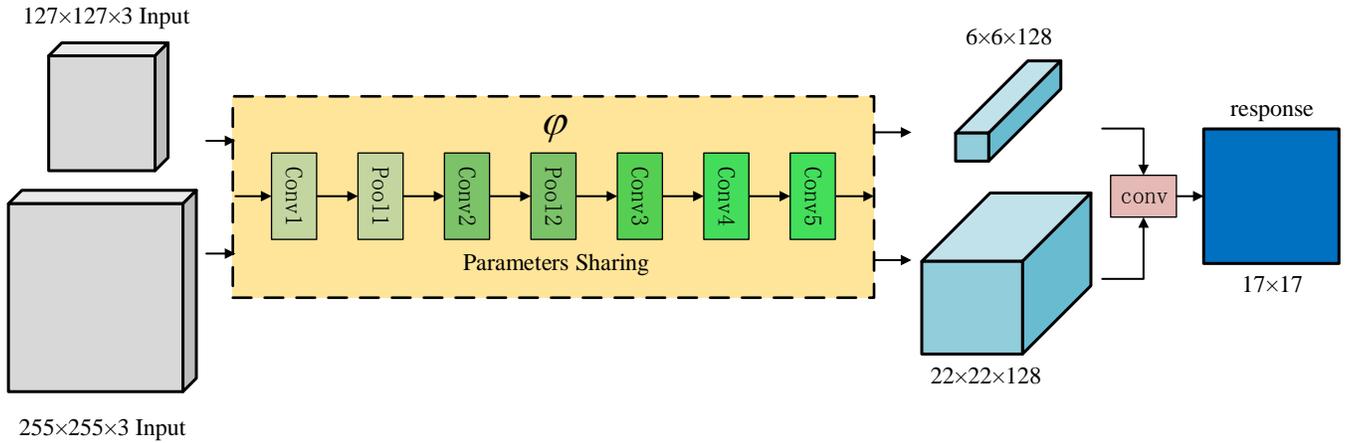

**Figure 5.** Bilinear network with parameters sharing. The two input with size of $255 \times 255 \times 3$ and $127 \times 127 \times 3$ pass through our bilinear network like AlexNet, and the size of final response will be $17 \times 17$. The parameters and feature dimensions of each layer is depicted in Table 1.

**Table 1.** Parameters and feature dimensions of SAS-Net

| Layer | Kernel size | stride | Size of attentional instance | Size of attentional exemplar |
|---|---|---|---|---|
| Input | | | $255 \times 255 \times 3$ | $* \times 127 \times 3$ or $127 \times * \times 3$ |

| Conv1 | 11 × 11 × 96 | 2 | 123 × 123 × 96 | ∗× 59 × 96 or 59 ×∗× 96 |
|---|---|---|---|---|
| Pool1 | 3 × 3 | 2 | 61 × 61 × 96 | ∗× 29 × 96 or 29 ×∗× 96 |
| Conv2 | 5 × 5 × 256 | 1 | 57 × 57 × 256 | ∗× 25 × 256 or 25 ×∗× 256 |
| Pool2 | 3 × 3 | 2 | 28 × 28 × 256 | ∗× 12 × 256 or 12 ×∗× 256 |
| Conv3 | 3 × 3 × 192 | 1 | 26 × 26 × 192 | ∗× 10 × 192 or 10 ×∗× 192 |
| Conv4 | 3 × 3 × 192 | 1 | 24 × 24 × 192 | ∗× 8 × 192 or 8 ×∗× 192 |
| Conv5 | 3 × 3 × 128 | 1 | 255 × 255 × 3 | ∗× 6 × 128 or 6 ×∗× 128 |

$$\begin{aligned} R^{att} &= \text{Sigmoid}\{conv[att(\varphi(Z_s)) \\ &\quad \odot \varphi(Z), att(\varphi(Z_s)) \odot \varphi(X)]\} \\ &= \text{Sigmoid}\{[att(\varphi(Z_s)) \odot \varphi(Z)] \\ &\quad * [att(\varphi(Z_s)) \odot \varphi(X)]\} \\ &= g(Z_s, Z, X; \theta_s, \theta_{att}) \end{aligned} \quad (2)$$

### Architecture of our bilinear network

The illustration of bilinear network structure is shown in Figure 5. The kernel size of each layer is similar to the AlexNet[13], the difference is that the fully convolutional structure is adopted in the network, and the fully connected layer is not included. The bilinear network is composed of layers of conv1, pool1, conv2, pool2, conv3, conv4, conv5. The rectified linear unit (RELU) is used after each convolution layer.

When the sizes of the input image patches are 127 × 127 × 3 and 255 × 255 × 3 respectively, the sizes of two features are respectively 6 × 6 × 128 and 22 × 22 × 128. Then the similarity response map can be generated by the convolution of the two features. The process can be described in formula (3).

$$\begin{aligned} R &= \text{conv}(\varphi(Z), \varphi(X)) \\ &= \varphi(Z) * \varphi(X) \\ &= f(Z, X; \theta_s) \end{aligned} \quad (3)$$

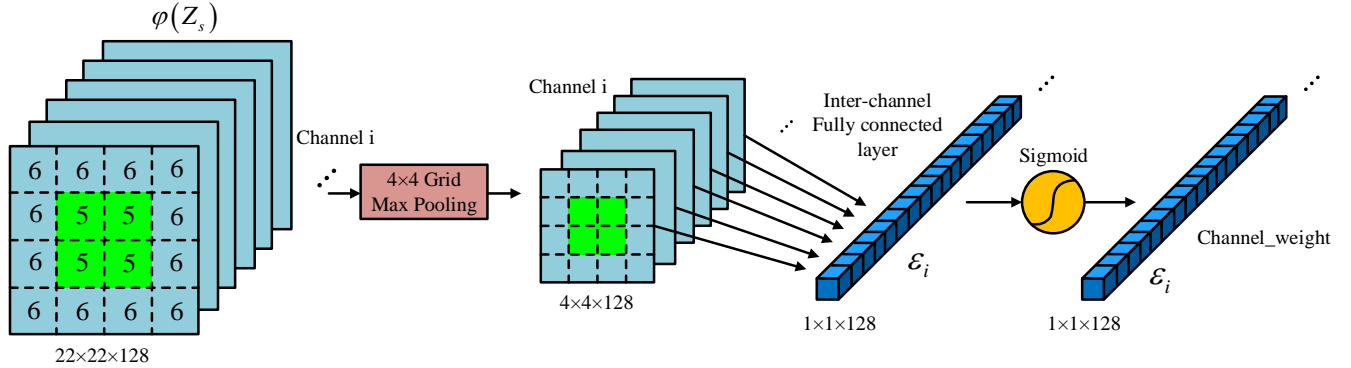

**Figure 6.** Inter-channel attention mechanism. The details of the attention model in Figure 5 is shown in this figure. The input is the semantic features obtained by bilinear networks. Through a process of the 4 × 4 grid max pooling and inter-channel fully connected layer, the final channel_weight will be generated.

The AlexNet does three times down-sampling operation with a stride of 2, apart from once convolution with a stride of 4. The overall scale is reduced by 31 times compared to the input image. The ∗ in the table indicates that the scale of the image or feature map is freely adaptive in this dimension.

Because the ratio of the length to width is kept during interpolation and the length of the long side is resized to 127 pixels, the length of the short side is adapt111e. During forward propagation of input exemplar, the output feature map size is computed according to the input feature size as follows:

$$\begin{aligned} h_{out} &= \lfloor(h_{in} + 2 \times h_{pad} - h_{kernel})/h_{stride}\rfloor + 1 \\ w_{out} &= \lfloor(w_{in} + 2 \times w_{pad} - w_{kernel})/w_{stride}\rfloor + 1 \end{aligned} \quad (4)$$

In the formula above, $h_{in}, w_{in}$ denote the height and width of input feature map and the input image especially in the first layer. Similarly, $h_{out}, w_{out}$ denote the height and width of input feature map. $h_{kernel}, w_{kernel}$ denote the height and width of convolution kernel or down-sampling kernel. $h_{stride}, w_{stride}$ denote the stride on height side and width side when the kernel is sliding on the feature maps. $h_{pad}, w_{pad}$ indicate the padding pixels on each side of feature maps if the input size can not divide the stride exactly. According to the formula above, the scale of the last feature map of the input exemplar $\varphi(Z)$ is $* \times 6 \times 128$, the scale range of $*$ is $[1,6]$. However, the size of the feature of input instance $\varphi(X)$ in the search area is fixed to $22 \times 22 \times 128$. The $response$ is obtained by the convolution of the exemplary feature $\varphi(Z)$ and the instance feature $\varphi(X)$, its size is $* \times 17$. The final convolution can be considered as the weighted sum calculated when the upper feature slides on the lower feature. The weighted sum is carried out in three dimensions: height, width and channels. The upper feature can be regarded as the convolution kernel. According to formula (4), outputSize $= \lfloor (22 + 2 \times 0 - 6)/1 \rfloor + 1 = 17$. In consideration of the adaptive dimension, the range of the length of corresponding side $*$ is $[17,22]$.

After the upsampling with step stride of 16, the range of the $*$ side of $response$ for positioning changes to $[272,352]$. Since all the transformation of the scale from the picture to the final feature map is consistent with the fixed input situation. The scaling ratio of the feature map has not changed.

The internal structure of the attention mechanism depicted in Figure 6 is the attention model in the main system model shown in Figure 4. The input is the semantic feature $\varphi(Z_s)$ output from the bilinear network in the first frame with the semantics $Z_s$ as the input. The output is the weight vector of each channel based on the semantic features. It is used to calculate the attentional exemplar and attentional instance in the main system. $\varphi(Z_s)$ is passed through a $4 \times 4$ grid max down-sampling layer. The specific implementations are as follows: First, the $22 \times 22$ feature map of each channel in $\varphi(Z_s)$ was divided into a $4 \times 4$ pixels grid, as shown in the left of Figure 6.

The 4 grids in the central area are $5 \times 5$ pixels size, as shown in the green region, and the sizes of the 12 white grids around are all $6 \times 6$ pixels; Then do max-pooling for all the pixels in each grid (as mentioned in the previous article), getting the $4 \times 4 \times 128$ feature maps after down-sampling. On the basis of the convolution principle and the extraction method of exemplar and background area, the features of the target vehicle should be located in the central area of the $\varphi(Z_s)$, as the green part of Figure 6 on the left. The target features combined with the surrounding pixels make up the semantic information of the attention mechanism. The $4 \times 4$ down-sampled feature maps of 128 channels are next passed through the inter-channel fully connected layer. The difference between inter-channel fully connected layer and

*Architecture of Inter-channel attention mechanism*

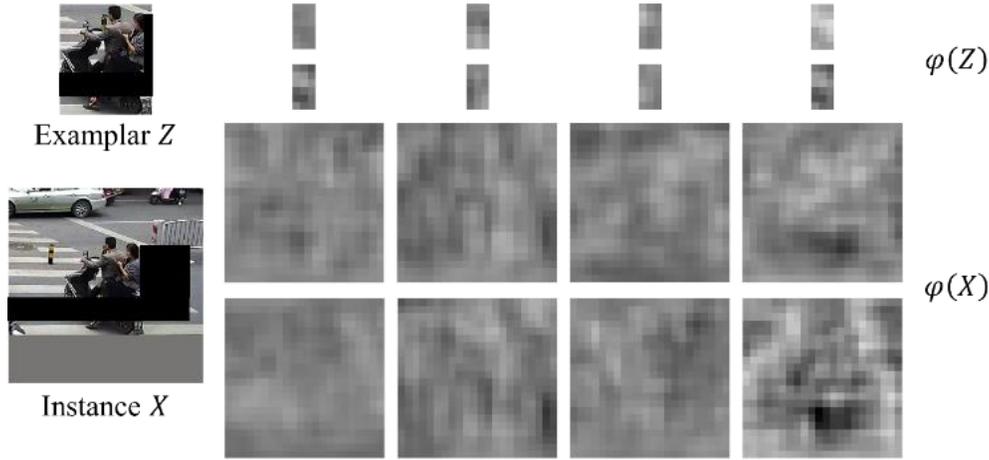

**Figure 8.** The first 8 layer feature maps of sample template and search area. The response of each channel feature to the target object is different. The convolution operation will pay more attention to the influence of the target object itself.

the ordinary fully connected layer is as follows: In the ordinary fully connected layer, all the pixels in the input layer are connected to all pixels in the output layer, it includes more parameters. If the feature size of the input layer is $4 \times 4 \times 128$ and the feature size of the output layer is $1 \times 1 \times 128$, the number of the parameters of this fully connected layer are $4 \times 4 \times 128 \times 128 = 262144$. The pixels in the input features of the inter-channel fully connected layer are only connected to the output vector in accordance with the corresponding channels. Namely, the i'th channel of the input feature is connected to the i'th channel of the output feature, so that there are only $4 \times 4 \times 128 = 2048$ parameters between the $4 \times 4 \times 128$ input feature and the $1 \times 1 \times 128$ output feature. The semantic features are mapped into a 128-dimensional feature vector after the inter-channel fully connected layer. Through a sigmoid function, all the values in the feature vector can be mapped to $[0,1]$, and the self-selected weights can be obtained. The weights based on the background semantics are multiplied by the exemplary feature $\varphi(Z)$ and the instance feature $\varphi(X)$ for each frame, and their self-selected feature maps can be obtained. In the model tracking stage, the same as the exemplar, it only needs to predict the weight vector of the feature maps of each channel according to the semantic information of the target vehicle and background provided by the first frame. No repeated calculation is needed in the subsequent frames.

*Extraction method of adaptive target scale*

In this paper, a target template extraction method based on the adaptive target size is proposed to break the rule that the network input should be fixed size.

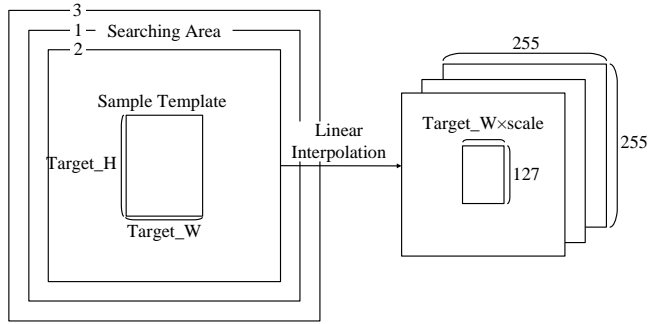

**Figure 7.** Adaptive sample template and search area. The long side of the target will be resized to 127 the searching area will be resized to $255 \times 255$, so do the other two scales.

To follow the target object, the target features can be adaptively changed, so the influence of the background to the target template can be reduced down to the lowest. Figure 7 shows a sample template without background area and the search area with the background filled around. The target box is directly used as the sample template with its long side interpolated to 127-pixel length. The scaling ratio of this procedure is calculated for interpolation of the short side at the same time.

In this way, the input of the network is no longer in a fixed size, but a rectangle with arbitrary aspect ratio. It is flexible for the network to adapt to various scales of the target vehicles, whether it is a moving motorcycle with larger aspect ratio or a driving car with relatively smaller aspect ratio. After determining the size of the sample template, a square search area is taken with the long side of target padded with a long side. The long edge of the rectangular target box is filled two times as long as before, and the short side is padded with more pixels from the background. Thus, when the search area is adjusted to 255×255 pixels, it can have the same scaling ratio with the target box and the sample template in the first frame.

Figure 8 is shown as the first eight layer feature maps of the exemplary feature $\varphi(Z)$ and the instance feature $\varphi(X)$. The input pair of variable scale based sample template $Z$ and search area $X$ is shown in the left. As introduced above, the exemplar $Z$ is almost cropped without background pixels in it. When feature $\varphi(X)$ is convoluted by feature $\varphi(Z)$, the number of feature points and multiplications involved in the operation is greatly reduced if the aspect ratio of the target vehicles increases. Compared to the circumstances in which a fixed square exemplar is input to the model, the tracking velocity has been improved. In the previous eight channels of features, it is obvious that the response of each channel feature to the target object is different. The convolution operation will pay more attention to the influence of the target object itself. The center area in the first and fifth channel features of $\varphi(X)$ does not seem to be very discriminable, while the apparent contours of the target are visible in fourth or eighth channels. Therefore, when the feature of the sample template and the feature of the search area are convoluted, the contribution to the similarity measurement of different channels should be different.

The channels that have a strong response of the target vehicles should be considered, and the channels which are not sensitive enough to the target object should be weakened. The attention mechanism is able to learn the importance of different channels effectively, make the model focus on the specific channels, increase effective calculation model. Based on the principle that different feature map channel has different importance, the fully convolutional neural network for tracking with a visual attention mechanism is proposed. The network can select different features according to the weight of features among multiple channels from the semantic patches of the input image.

## Loss Function and Training

The parameters of our inter-channel attention based fully convolutional neural tracking model are mainly divided into two parts, one part is the parameters of the convolution kernels of the fully convolutional layer in the shared bilinear network, the other part is the parameters involved in the inter-channel fully connected layer in the visual attention mechanism. So a two-stage method for training our model is adopted. First we use a large public dataset to pre-train bilinear network, making the bilinear part achieve an optimal result on this dataset. Then use our traffic dataset to train the entire network including the visual attention part, while fine-tuning the parameters of the bilinear part. A large learning rate is used for the attention mechanism, and a lower learning rate is used for the bilinear network. At the same time, the parameters of the whole model are optimized jointly.

According to the section "Similarity neural network", the method of labeling the similarity of an image block with bool value 0 and 1 is too "absolute". In this way, the intermediate state is never involved. The judgment of the similarity should be an analog process, but the 0-1 annotations digitize the probability distribution. This will inevitably result in the loss of information. In the proposed method, we use an analog tagging method to train the network, and finally achieve good results.

When the bilinear network is trained in the first stage, the network input is the image patch pair $(Z, X)$ composed of the sample template Z and the search area $X$. The extraction method of them in the original image frames has been introduced in the previous article. Because the size of the input template is adaptive to the size of the target in input image, the size of the similarity response map is $* \times 17$, $* \in [17,22]$, and the two-dimensional gaussian label matrix Y can be generated at the center of the response map, as shown in Figure 9.

**Figure 9.** Similarity response and two-dimensional label. Different from traditional binary labels to each pixel, an analogue label method is applied.

Because the image patch pair $(Z, X)$ is in the center of the search area when the image patch is extracted, the meaning of the label matrix is that the closer distance to the center of the response map, the closer distance to the target template, the higher value the similarity is. On the contrary, the farther away from the center, the farther the distance from the target template, the lower value the similarity will be greatly reduced. The loss function at the location of $(i,j)$ is defined as:

$$l(i,j) = \log\left(1 + \exp(-Y(i,j)R(i,j))\right) \qquad (5)$$

where $R(i,j)$ is the response value at the position $(i,j)$ in the output response map, and $Y(i,j)$ is the value of the two-dimensional label matrix at the position $(i,j)$. If the similarity response value at the location $(i,j)$ and the two-dimensional label matrix value $Y(i,j)$ are both high, the similarity degree of the sample template at the position $(i,j)$ and this search area is higher than that in other search areas. Obviously, $l(i,j)$ is a monotonically decreasing function of $Y(i,j)R(i,j)$. The position of higher similarity has a lower loss value. Then, with $n'$th pair of the sample template and the search area as the network input, the overall loss function of the similarity response map is defined as:

$$L(f(Z_n, X_n; \theta_s), Y_n) = \frac{1}{|D_n|} \sum_{i,j \in D_n} l_n(i,j)$$
$$= \frac{1}{|D_n|} \sum_{i,j \in D_n} \log\left(1 + \exp(-Y_n(i,j)R_n(i,j))\right) \quad (6)$$

where $(Z_n, X_n)$ is the $n'$th pair of the sample template and the search area. $D_n$ represents the coordinates set of output response map with $n'$th pair image patch as input. $R_n$ and $Y_n$ represent the $n'$th output response map and the $n'$th two-dimensional label matrix according to the size of $R_n$. $l_n(i,j)$ is the loss at position $(i,j)$ on the $n'$th output response map $R_n$ defined in formula (5). $\theta_s$ represent all the parameters in similarity bilinear network.

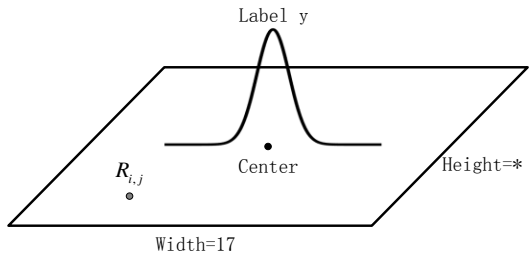

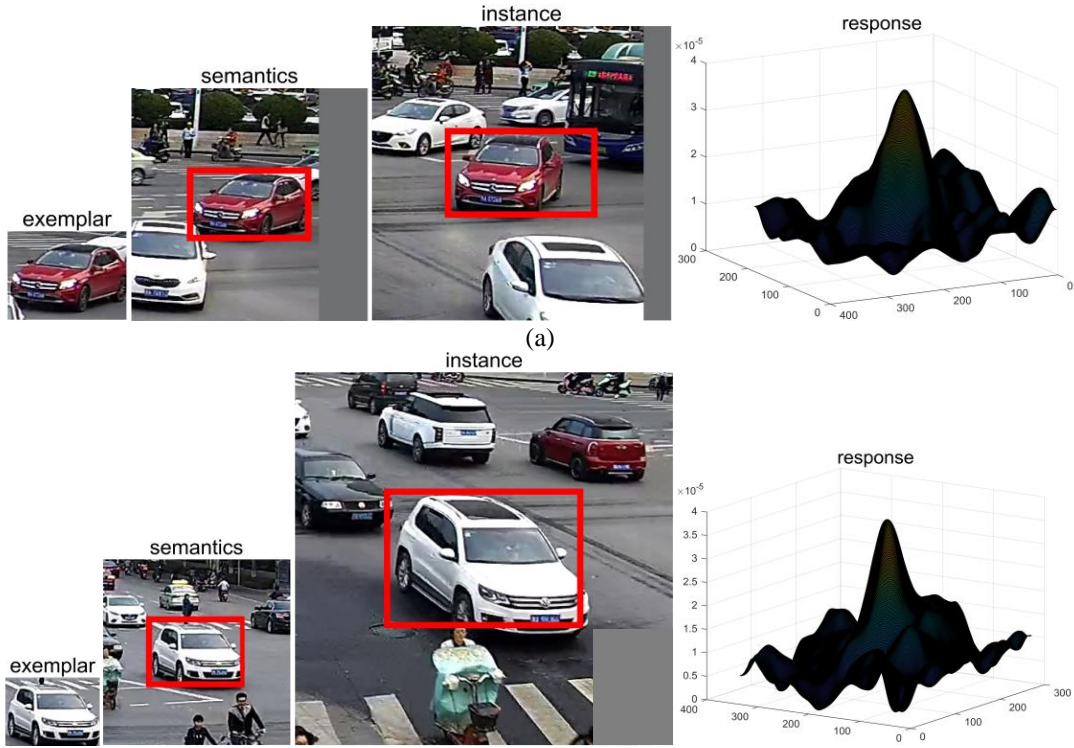

(a)

(b)

**Figure 10.** Exemplar, semantics, instance and response map from traffic dataset. The rectangular box of sample template tightly encapsulates the target vehicle, the semantic feature of the background area and the feature of the search area extracted around the rectangular target box in the process of tracking. All of them are interpolated according to the target scale.

During training, the Batch Gradient Descent (BGD) is used to optimize the parameters of the bilinear part, and the whole process can be written as an optimization function:

$$\underset{\theta_s}{\arg\min} \, \mathbb{E}[L(f(Z_n, X_n; \theta_s), Y_n)] \quad (7)$$

where $\mathbb{E}$ represents the mathematical expectation of the loss function of all sample pairs. After calculating the loss function value of each sample through the network, BGD is used to find and update the partial derivative of parameters $\theta_s$ in the bilinear network by backpropagation. Thus the optimal parameter for the mathematical expectation of the loss function is finally obtained. In this procedure, the learning rate of the optimizer is set to 0.001. The number of batch pairs is 4 during training. The traffic vehicle dataset in our laboratory is not quite sufficient compared with some public datasets, and the neural network built in this paper is large scale.

This paper chooses to use the public dataset based on ImageNet Large Scale Visual Recognition Challenge (ILSVRC) for pre-training bilinear network to prevent the overfitting problem of the model in the pre-training stage. The dataset contains about 4500 video sequences and 30 different vehicles and animals. Compared to other datasets, the target species and the number of video sequences in the dataset satisfy the requirements of proposed network in this article.

After the pre-training stage of our bilinear network is completed using the ILSVRC dataset, the traffic dataset in this article is used to train the visual attention mechanism and fine-tune the parameters in the bilinear network. The traffic dataset consists of 40 video sequences, which contain crossroads, bus stations, and other densely populated scenes. The targets in the video are mainly traffic participants, such as vehicles, bicycles, and pedestrians. Figure 10 shows the sample template, the semantic information and the search area extraction method under the traffic dataset of our laboratory. The rectangular box of sample template tightly encapsulates the target vehicle, the semantic feature of the background area and the feature of the search area extracted around the rectangular target box in the process of tracking. All of them are interpolated according to the target scale. Sample template and semantic information are extracted from the initial frame of the video, and the search area is extracted in each subsequent frame. When the sample template is input into the network, it is adjusted to $* \times 127$ or $127 \times *$ size, semantic information and search area are adjusted to $255 \times 255$ size.

In the second stage of joint training, the loss function at the location $(i,j)$ of response map $R^{att}$ is still defined according to formula (8):

$$l(i,j) = \log\left(1 + \exp(-Y_{i,j} R^{att}_{i,j})\right) \quad (8)$$

where $R^{att}_{i,j}$ is the value at location $(i,j)$ on the ouput similarity map of SAS-Net, $Y_{i,j}$ is the label value at location $(i,j)$ as defined before. Then, with $n'$th pair of the exemplar and the instance as the network input, the overall loss function of the similarity response map of SAS-Net is:

$$L(g(Z_s, Z_n, X_n; \theta_s, \theta_{att}), Y_n)$$
$$= \frac{1}{|D_n|} \sum_{i,j \in D_n} l_n(i,j) \quad (9)$$
$$= \frac{1}{|D_n|} \sum_{i,j \in D_n} \log\left(1 + \exp(-Y_n(i,j) R^{att}_n(i,j))\right)$$

where $(Z_s, Z_n, X_n)$ is the $n'$th combination of the semantics, the exemplar and the instance, $R^{att}_n$ represents the similarity response of SAS-Net after adding the visual attention mechanism. $\theta_s$ and $\theta_{att}$ represent the parameters of similarity bilinear network and visual attention mechanism respectively.

In the second training stage, we use the RMSprop method to optimize the network parameters $\theta_s$ and $\theta_{att}$ jointly. This is a very effective optimizer for improving the adaptive learning rate. It will no longer sum up all the squares of the gradient during training, but reduce it by an attenuation rate. It uses a moving average method, by which the influence of the gradient of previous training samples on the adaptive learning rate is smaller. It can avoid the problem in other optimizers that the learning rate has been reduced too much, and can converge faster. The whole process can be written as an optimized function:

$$\underset{\theta_s,\theta_{att,}}{\operatorname{argmin}} \mathbb{E}[L(g(Z_s, Z_n, X_n; \theta_s, \theta_{att}), Y_n)] \quad (10)$$

The entire network uses different initial learning rates for $\theta_s$ and $\theta_{att}$ during training (0.0001 and 0.001 respectively). Because the parameter $\theta_s$ has been pre-trained on the ILSVRC dataset, we only need a small learning rate for $\theta_s$ and a normal learning rate for $\theta_{att}$.

There is no difference between the forward propagation of the network in the test tracking stage and the training stage. The sample feature $\varphi(Z)$ and the semantic feature $\varphi(Z_s)$ can be obtained at the initialization stage. The weight value Channel_weight of the multi-channel features is calculated by the attention mechanism with the semantic features as the input. In tracking process of each subsequent frame, only the multi-scale instance $X$ need to be repeatedly extracted to compute the branch of $\varphi(X)$ and Search_attended feature map.

## Experiments

The experimental environment for this paper is as follows: CPU Intel core i7-4790 3.6GHz; memory 16GB, GPU NVIDIA Quadro K2200, which contains 640 CUDA computing core units and 4GB graphics memory. The simulation software is matlab R2017a, and the program is written mainly based on the deep learning framework Matconvnet for Matlab. The evaluation video data is provided by the intelligent traffic big data project of our laboratory. 40 video sequences are truncated, and the lengths of the truncated videos are not equal. The frame rate of these videos is 25 frames per second (FPS), and the resolution is 1920 × 1080. It includes the traffic scene like bus stations, intersections, crosswalks and other people intensive areas. The conditions of the road are extremely complex. The performance comparison between SAS-Net, Siamese network, and KCF, DSST methods based on correlation filters on our traffic dataset in this paper is shown in the Figure 11.

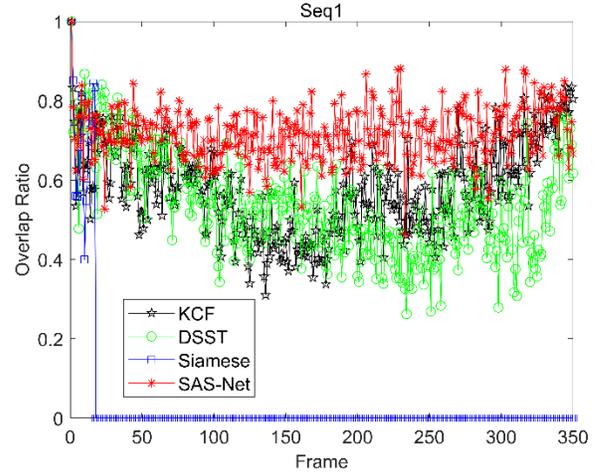
(a)

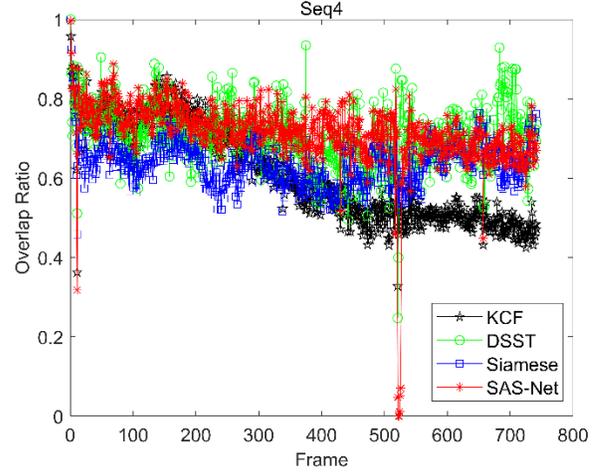
(b)

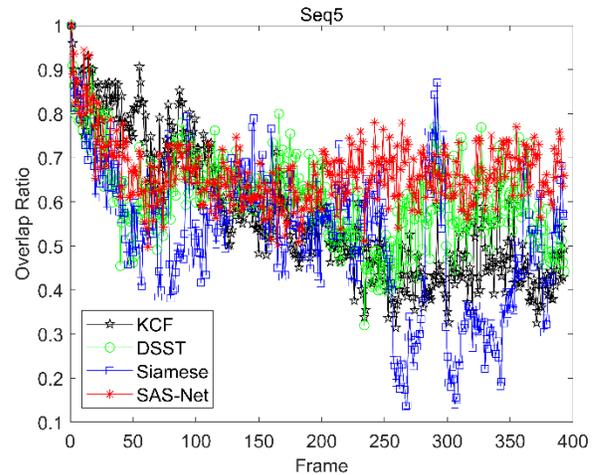
(c)

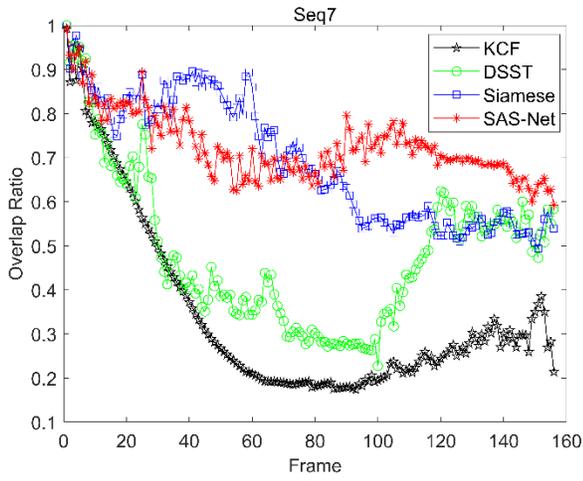

(d)

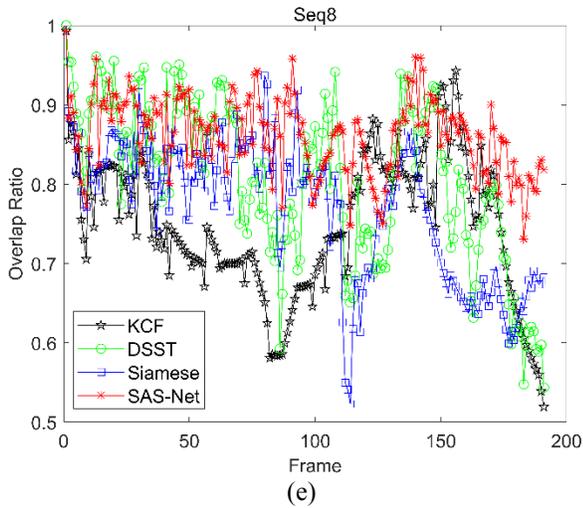

(e)

**Figure 11.** Overlap ratio curves of several video sequences. Overlap ratio is the ratio of the overlapping area to the merging area between predicted box and groundtruth box.

Figure 11 shows overlap ratio curve of each frame in sequence 1, sequence 4, sequence 5, sequence 7 and sequence 8 in our traffic dataset. In sequence 1, the Siamese network completely loses tracking of the target vehicle at about 20 frames because of occlusion or background interference, while SAS-Net performs well. The overall overlap ratio curve has been higher than the curves of the other three methods. In sequence 4, SAS-Net and DSST have a fairly high overlap ratio. Both the overlap ratio curves show that the overall overlap ratio between the predicted rectangular box and the real vehicle rectangular box is higher than Siamese when the SAS-Net is tracking. Only the overlap ratio of several frames is lower than the Siamese network. In sequence 5, before the 150 frame, the overlap ratio of SAS-Net is only inferior to that of KCF. After 150 frames, the overlap ratio of SAS-Net is higher than them of the other three methods. In sequence 7, the overlap ratio of SAS-Net is higher than that of KCF and DSST, it is also higher than Siamese after 80 frames. The overlap ratio of SAS-Net in sequence 8 is higher than that of other three methods.

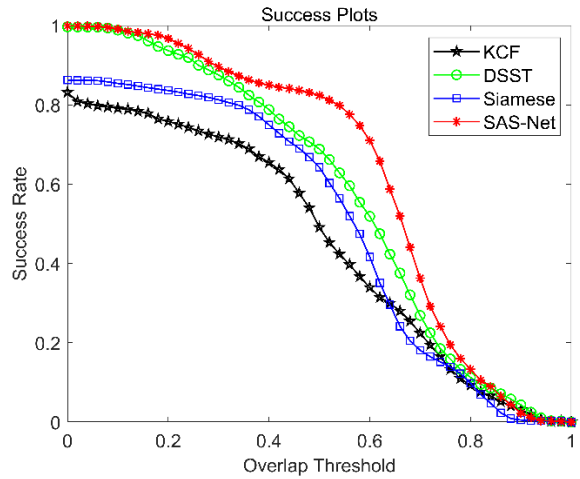

**Figure 12.** Success rate with the overlap ratio threshold curve. The success rate is defined as the ratio of the success number to the total number of frames in all videos. The success number is the sum of success frames where the overlap ratio between the predicted box and the ground-truth box is higher than a certain threshold.

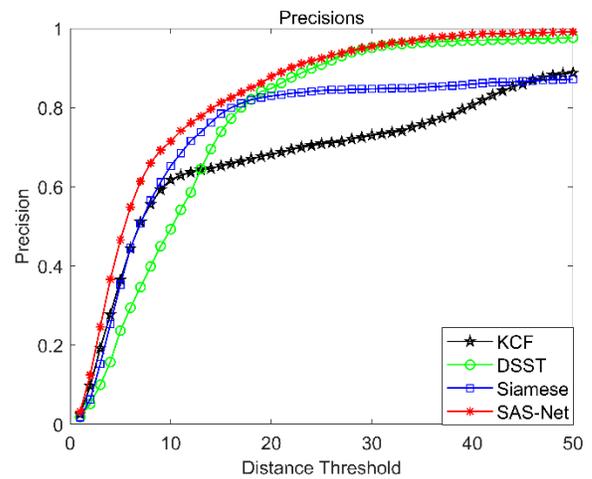

**Figure 13.** Precisions with the distance threshold curve. The points on the curve is the ratio of the number of accurate frame number to the number of all testing video frames. The accurate frame is a frame where the distance between a predicted rectangular box and a ground-truth box is lower than a certain distance threshold on the abscissa axis.

Figure 12 shows the success rate and the thresholds of overlap ratio in 20 sequences of SAS-Net, Siamese, KCF, and DSST. The success rate is defined as the ratio of the success number to the total number of frames in all videos. The success number is the sum of success frames where the overlap ratio between the predicted box and the ground-truth box is higher than a certain threshold. Overlap ratio is the ratio of the overlapping area to the merging area between two rectangular boxes. It can be seen from the figure that the success rate of SAS-Net tracking is higher than that of other three methods under most overlap ratio thresholds.

Figure 13 shows the precisions with distance threshold curves of SAS-Net, Siamese and correlation filters methods of KCF and DSST in the 20 sequences. The meaning of the points on the curve is the ratio of the number of accurate frame number to the number of all testing video frames. The accurate frame is a frame where the distance between a predicted rectangular box and a ground-truth box is lower than a certain distance threshold on the abscissa axis. The distance is calculated by the Euclidean distance between the two central points of two rectangular boxes. As can be seen from the figure, the precision of SAS-Net is much higher than the other three methods when the distance threshold is below 30. When the distance threshold is higher than 30, the precision of SAS-Net is not the same as that of the DSST method, and is still much higher than the KCF and Siamese methods by 10 percent. Compared with the Siamese network, the SAS-Net has greatly improved its performance.

Figure 14 shows the tracking results of SAS-Net, Siamese, KCF and DSST in several key frames of our traffic video sequences. It can be seen from the figure that when the target vehicle is far away from the camera, the sensor cannot capture the features of the vehicle better. This will lead to the drift of the model so that the KCF fail to follow the target vehicle. Compared with several other methods, the Siamese method has a poor tracking result when the target vehicle is in a complex background area. The main reason is that the deep features of the surrounding vehicles are too similar to the deep features of the target vehicles, but the influence of surrounding background information to the model is not considered when the sample template is extracted. In this paper, the SAS-Net method effectively filters the deep features at different channels through the attention mechanism. It multiplies each channel by different coefficients according to the importance of different channels. The vehicles are no longer disturbed by the background area and the occluded vehicles. Our model has achieved good results compared with other methods.

Table 2 shows a contrast between SAS-Net and Siamese networks, KCF and DSST. Evidently, the average processing speed of the Siamese network on 20 video sequences is 22.98 FPS, while the average processing speed of the SAS-Net on the 20 video sequences is 28.51FPS, slightly faster than the Siamese network. This is mainly due to the variable scale sample template adopted in this paper. The number of convolution features calculated by similarity measurement is greatly reduced, so the speed is improved when the last two features are convoluted. The speed of the SAS-Net is also faster than DSST by over 10 FPS. This is because the DSST method also trains a scale filter online while updating the location filter, and extracts the target features of about 30 times when testing and updating. Without the support of the GPU lab environment, this process is time consuming. However, SAS-Net only needs to extract the features of three times in the calculation, and can achieve the desired speed with the support of Matconvet framework. So the two methods also apply to different scenarios. Table 3 exactly shows the comparison on time for the convolution of sample template feature $\varphi(Z)$ and search area features $\varphi(X)$. SAS-Net saves an average of 0.004s time in the final convolution process compared with the Siamese network. As $\varphi(X)$ is convoluted by $\varphi(Z)$, the feature size is reduced, so the number of multiplication operations made by convolution is also greatly reduced.

**Table 2.** FPS of SAS-Net compared with other methods

| Methods | Tracking speed on 20 video sequences (FPS) | | | | | Average |
|---|---|---|---|---|---|---|
| KCF | 208.79 | 113.21 | 107.08 | 216.89 | 61.65 | 145.51 |
| | 176.41 | 150.54 | 141.38 | 143.91 | 97.23 | |
| | 179.83 | 152.29 | 261.97 | 199.52 | 102.41 | |
| | 221.44 | 82.42 | 47.29 | 218.09 | 27.82 | |
| DSST | 13 | 26.7 | 26.1 | 34.3 | 12.4 | 16.55 |
| | 32.1 | 24.2 | 22.4 | 27.1 | 5.9 | |
| | 29.8 | 9.2 | 13 | 11 | 7.58 | |
| | 13.1 | 7.23 | 2.11 | 11.9 | 1.9 | |
| | 63.19 | 68.22 | 55.61 | 53.42 | 28.81 | |
| | 85.44 | 82.18 | 102.16 | 72.68 | 48.20 | |
|         | 93.89 | 31.12 | 3.79  | 68.75 | 2.25  |       |
| Siamese | 23.68 | 23.99 | 24.04 | 24.54 | 21.84 | 22.98 |
|         | 23.84 | 22.36 | 23.26 | 23.44 | 23.79 |       |
|         | 23.44 | 23.32 | 22.18 | 24.31 | 22.64 |       |
|         | 22.43 | 22.63 | 19.03 | 23.07 | 21.72 |       |
| **SAS-Net** | **29.21** | **29.28** | **29.64** | **29.98** | **27.83** | **28.51** |
|         | **29.01** | **28.55** | **29.45** | **29.57** | **29.69** |       |
|         | **29.41** | **29.85** | **28.57** | **27.48** | **28.22** |       |
|         | **28.09** | **28.26** | **25.34** | **24.96** | **27.97** |       |

**Table 3.** Comparison of consumption time for feature convolution

| time of convolution between $\varphi(Z)$ and $\varphi(X)$ (seconds) ||||
|---|---|---|---|
| Siamese | 0.01 | **SAS-Net** | **0.006** |

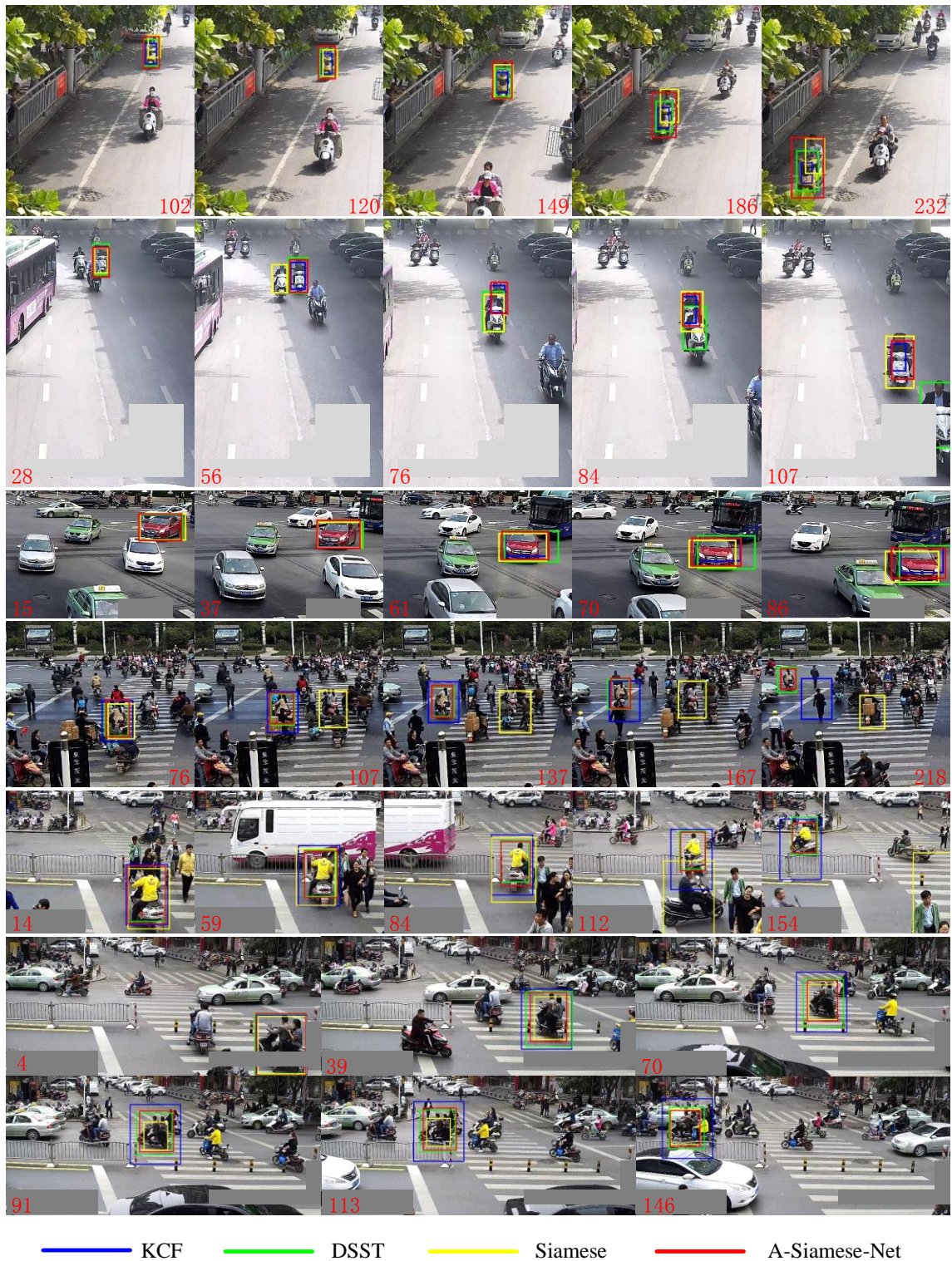

**Figure 14.** Tracking results on key frames in part of video sequences. When the target vehicle is far away from the camera, the KCF method cannot capture the features of the vehicle better. The Siamese method has a poor tracking result when the target vehicle is in a complex background area. Our SAS-Net has achieved good results compared with other methods.

## Conclusion and future work

In this article, we focus on the vehicle tracking task for IOV and propose the SAS-Net model which combines the bilinear network with a visual attention mechanism, making the model select the features from different channels according to different impacts of features from semantic areas. The model makes the vehicle no longer disturbed by the background area and the occluded vehicles, and has obtained good results compared with other methods. The success rate of SAS-Net tracking is higher than that of KCF, DSST and Siamese-FC under most overlapping rate thresholds. The precision of SAS-Net is much higher than the other three methods when the distance threshold is below 30. Our SAS-Net achieves a real-time tracking speed. If there is more hardware support, this model can be applied to large-scale intelligent IOV system to assist all the connected vehicles to make a good travel plan.

In the future research work, the joint vehicle tracking method based on the multi-image sensor will be brilliant. In the IOV system, interaction and fusion of information generated by different sensors among different vehicles are inevitable. In the combination of multi-sensors, how to fuse the multimodal data from various kind of sensors, such as image sensors and wireless sensors, will be a long-term research goal. Deep neural networks still have great potential in these fields. Designing different kinds of neural networks for these multimodal data can achieve the fusion of different sensor information.